\documentclass{article} % For LaTeX2e
\usepackage[table]{xcolor}
\usepackage{iclr2023_conference,times}

\usepackage{amsmath,amsfonts,bm}

\def\eqref#1{equation~\ref{#1}}
\def\1{\bm{1}}

\DeclareMathAlphabet{\mathsfit}{\encodingdefault}{\sfdefault}{m}{sl}
\SetMathAlphabet{\mathsfit}{bold}{\encodingdefault}{\sfdefault}{bx}{n}

\usepackage[hidelinks]{hyperref}
\usepackage{url}

\usepackage{arydshln}

\usepackage{wrapfig}
\usepackage{booktabs}
\usepackage{pifont} 
\newcommand{\cmark}{\ding{51}}%
\newcommand{\xmark}{\ding{55}}%

\usepackage{xspace}
\usepackage{graphicx, amsmath, amssymb, caption, subcaption, multirow, overpic, textpos}
\usepackage{wrapfig}
\usepackage{tabulary}
\usepackage[british, english, american]{babel}

\newlength\savewidth\newcommand\shline{\noalign{\global\savewidth\arrayrulewidth
		\global\arrayrulewidth 1pt}\hline\noalign{\global\arrayrulewidth\savewidth}}
\newcommand{\tablestyle}[2]{\setlength{\tabcolsep}{#1}\renewcommand{\arraystretch}{#2}\centering\footnotesize}
\renewcommand{\paragraph}[1]{\vspace{1.25mm}\noindent\textbf{#1}}

\newcolumntype{x}[1]{>{\centering\arraybackslash}p{#1pt}}
\newcolumntype{y}[1]{>{\raggedright\arraybackslash}p{#1pt}}
\newcolumntype{z}[1]{>{\raggedleft\arraybackslash}p{#1pt}}

\newcommand{\app}{\raise.17ex\hbox{$\scriptstyle\sim$}}

\newcommand{\x}{{\times}}
\definecolor{deemph}{gray}{0.6}

\definecolor{baselinecolor}{gray}{.9}
\newcommand{\baseline}[1]{\cellcolor{baselinecolor}{#1}}
\def\x{$\times$}

\newcommand{\tblref}[1]{Table~\ref{#1}}

\usepackage{graphicx}
\title{DecoupleVideo: Decoupling Spatiotemporal Prediction for Self-Supervised Video Representation Learning}

\title{It takes two: Motion Synergized for Pre-training of Video Transformers}

\title{VideoCAE: Context Autoencoder for Self-Supervised Video Representation Learning}

\title{It Takes Two: Masked Appearance-Motion Modeling for Self-supervised Video Transformer Pre-training}

\author{Yuxin Song$^{*}$, Min Yang\thanks{Equal contribution, \{songyuxin02 and yangmin09\}@baidu.com}, Wenhao Wu, Dongliang He\thanks{Corresponding author. Email: hedlcc@126.com}, Fu Li, Jingdong Wang \\
Department of Computer Vision (VIS) Technology, Baidu Inc.\\
}

\iclrfinalcopy % Uncomment for camera-ready version, but NOT for submission.
\begin{document}

\maketitle

\begin{abstract}

Self-supervised video transformer pre-training has recently benefited from the mask-and-predict pipeline. They have demonstrated outstanding effectiveness on downstream video tasks and superior data efficiency on small datasets. However, temporal relation is not fully exploited by these methods. 
% Solution
In this work, we explicitly investigate motion cues in videos as extra prediction target and propose our \textbf{M}asked \textbf{A}ppearance-\textbf{M}otion \textbf{M}odeling (MAM$^2$) framework. 
Specifically, we design an encoder-regressor-decoder pipeline for this task. The regressor separates feature encoding and pretext tasks completion, such that the feature extraction process is completed adequately by the encoder. In order to guide the encoder to fully excavate spatial-temporal features, two separate decoders are used for two pretext tasks of disentangled appearance and motion prediction. 
% advantages
We explore various motion prediction targets and figure out RGB-difference is simple yet effective. As for appearance prediction, VQGAN codes are leveraged as prediction target.
% We find simple RGB cues across video frames are the best for motion decoder.(reason?). 
With our pre-training pipeline, convergence can be remarkably speed up, \emph{e.g.,} we only require half of epochs than state-of-the-art VideoMAE (400 \emph{v.s.} 800) to achieve the competitive performance.
% Exp results
Extensive experimental results prove that our method learns generalized video representations. Notably, our MAM$^2$ with ViT-B achieves 82.3\% on Kinects-400, 71.3\% on Something-Something V2, 91.5\% on UCF101, and 62.5\% on HMDB51. 

\end{abstract}
\section{Introduction}
%%% rethink the disentangled story
% background
With regard to a variety of video tasks, the Vision Transformer \citep{dosovitskiy2020vit} has achieved impressive results in supervised and unsupervised settings.
For supervised learning, ViViT \citep{arnab2021vivit}, TimeSformer \citep{bertasius2021timesformer}, Video Swin \citep{liu2021videoswin}, and MViT \citep{fan2021mvit} are emerging and achieve favorable performance in video understanding related tasks by using labeled data. 
On the other hand, contrastive learning based self-supervised methods, e.g., MoCo \citep{he2020momentum}, SimCLR \citep{chen2020simple} and BYOL \citep{grill2020bootstrap} etc., have also been widely extended to video representation learning. In \citep{feichtenhofer2021largescale}, a large-scale study on spatial-temporal representation learning using transformers and convolutional neural networks are conducted. 
%It inspires robust video representation could be learned in video-level temporal range
More recently, a series of methods which explore masked video modeling \citep{wang2022bevt, tong2022videomae, feichtenhofer2022videomaefb} have been proven to be promising for self-supervised representation learning.
Specifically, the \textit{mask-and-predict} pipeline of MAE \citep{he2021mae} is extended to the video domain by ST-MAE \citep{feichtenhofer2022videomaefb} and VideoMAE \citep{tong2022videomae}, which reconstruct the raw pixels of masked video patches from the visible contexts in a spacetime-agnostic manner. 
Besides, BEVT \citep{wang2022bevt} jointly reconstructs the discrete visual tokens obtained by an offline trained tokenizer in the image and video domain. 

% problem
However, the information density of video is much lower than image. Although ST-MAE \citep{feichtenhofer2022videomaefb} and VideoMAE \citep{tong2022videomae} employ a high-ratio masking strategy to reduce the redundancy in the Spatio-temporal domain and holistically comprehend the video beyond low-level visual statistics, they are still constrained in their visual contents and suffer from the inconsistency between the masked pre-training with space-aware target and the video understanding task. 
To the best of our knowledge, no research has previously been done on how to leverage potential motion clues in video to guide pre-training in the \textit{masked video modeling} manner.
% On the other hand, ViViT \citep{arnab2021vivit} and TimeSformer \citep{bertasius2021timesformer} factorize the joint space-time attention into a spatial and temporal dimension on the representation level and have achieved excellent performance in video representation learning.

% motivation & solution
Based on the aforementioned points, we creatively use two disentangled decoders to simultaneously reconstruct the visual appearance and motion targets.
Specifically, we first study on the various motion cues that are concealed in the video data, such as optical flow, RGB difference, and temporal order \citep{xu2019self}.
Then we employ visual tokens generated by the {\color{black}{discrete variational autoencoder (dVAE) \citep{esser2021taming, van2017neural} as a target}} for visual reconstruction that offers supplementary semantic clue to motion information during the  pre-training phase. 
Intuitively, we design a pair of disentangled decoders, one to reconstruct the visual tokens and the other to reconstruct the motion target.
By forcing the encoder to learn inherent spatio-temporal relation in video data, two independent disentangled decoders enable the encoder to be more effectively transferred to downstream tasks. 
Additionally, decoupling the appearance and motion view on the target-level allows for the efficient handling of redundant spatio-temporal data, speeding up convergence in the pre-training phase as shown in Fig.\ref{fig:compare}.

To eliminate the discrepancy between the view of appearance and motion, we adopt a regressor in \citep{chen2022context} to further project the latent representation from the encoder in a spacetime-agnostic manner. In more detail, the regressor learns to query from embeddings of the visible tubes and predict the representations for masked video tubes. These predicted representations of masked tubes are constrained to be aligned with their groundtruth encoding output and are then utilized for completing pretext tasks. With the regressor, the feature encoding and pretext tasks completion are explicitly separated. Therefore, our MAM$^2$ prevents the encoder from being entangled in different objectives of reconstruction during pre-training, guiding the encoder to focus on spatial-temporal representation learning and fully excavate spatial-temporal features.

%fig1.
%k400, train-poech, flops, acc

% contribution
With our MAM$^2$ pre-training framework, we empirically show that convergence can be remarkably speed up and the learned video representations generalize very well on multiple standard video recognition benchmarks.
In a nutshell, our contributions are as follows:
\begin{itemize}
\item We decouple the visual appearance and motion views to focus on different target in the vein of self-supervised video pre-training. As far as we are aware, we are the first to explore practical and efficient motion cues in the \emph{Masked Video Modeling} pre-training paradigm.

\item We propose a novel Masked Appearance-Motion Modeling (MAM$^2$) framework which reconstructs the appearance and motion targets separately. MAM$^2$ learns generalized spatio-temporal representation which better benefits the downstream video tasks. Additionally, we adopt a spacetime-agnostic regressor to avoid the encoder being entangled with pretext tasks, thus focusing on spatial-temporal feature excavation.

\item Extensive investigation results on standard video benchmarks demonstrate that our MAM$^2$ achieves a significant performance improvement than previous state-of-the-art methods with even half of pre-training epochs. Specifically, our MAM$^2$ brings the gain of 1.0\% on Kinetics-400, 0.7\% on Something-Something V2 with a 2$\times$ pre-training epochs speedup, and obtains the improvement of 0.7\% on UCF101 and 1.4\% on HMDB51 with a 3$\times$ pre-training epochs speedup.%, as shown in \textcolor{red}{(Fig.1)}.

\end{itemize}

%1. Dual decoder, decouple the visual semantics and temporal motion prediction. 
%2. The latent contextual regressor makes predictions in the latent space from the visible patches to the masked patches.
%3. sota, less pretraining epoch

%\textcolor{blue}{replace n.: 
%VideoCAE, semantic decoder, motion decoder, decoupling decoders, coupling decoder, alignment constraint}

\section{Approach}
\begin{figure*}[!t]
  \centering
  \includegraphics[width=\textwidth]{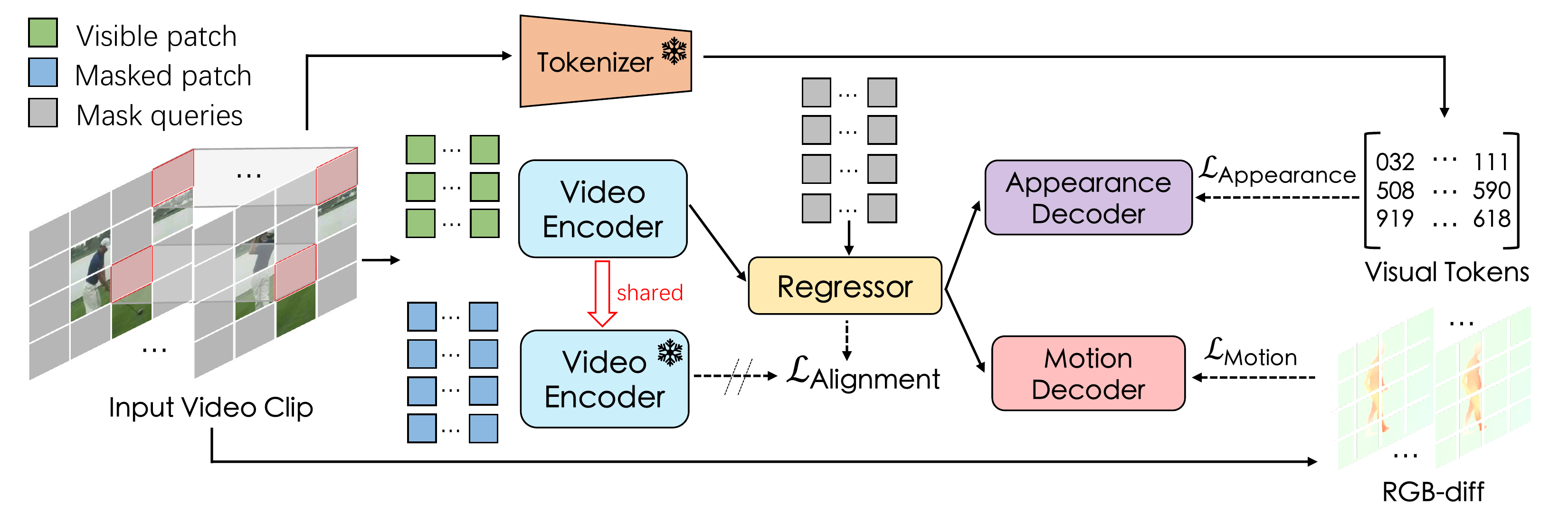}
  \caption{Overall architecture of our MAM$^2$ pretraining framework, which is an encoder-regressor-decoder pipeline. 
  %Input video clip is divided into visible(in green) and masked patches(in blue) by a random \emph{tube} masking strategy. The visible and masked patches are injected into the encoder and a copied one(frozen in training) which shares weights with the other respectively to generate visible and masked embeddings. Learnable mask queries(in gray) are used to predict the latent representations of masked patches by the regressor. Meanwhile, an alignment module is applied to pull the latent representations and masked embeddings close during training. Dual-stream disentangled decoders are designed to predict visual tokens and motion targets of raw video.
  }
  \vspace{-5mm}
  \label{fig:overview}
\end{figure*}
We propose the Masked Appearance-Motion Modeling (MAM$^2$) framework for self-supervised video representation learning. 
Our MAM$^2$ equips with the encoder-regressor-decoder architecture to prevent the encoder from becoming involved in the reconstruction task.
To excavate the spatial-temporal features in the video stream, we establish two distinct pretext tasks to independently decode the visual and motion information of video data to guide the encoder learning process. 
In Sec.~\ref{sec:architecture}, we describe the architecture design of our MAM$^2$. We then go over several potential motion objectives for pre-training before introducing the optimization function in Sec.~\ref{sec:targets}.

%\subsection{Overview of MAMM}
\subsection{Architecture}
\label{sec:architecture}
The architecture of our method is depicted in Fig.\ref{fig:overview}. The encoder is pre-trained for video representation learning. A latent contextual regressor learns to query from embeddings of the visible tubes
and predict the representations for masked video tubes. These predicted representations of masked
tubes are constrained to be aligned with their groundtruth encoding output and are then utilized
for completing pretext tasks. Dual-stream disentangled decoders are leveraged for reconstructing appearance and motion pre-training targets.

\textbf{Input.} 
A video clip $\mathbf{X}\in \mathbb{R}^{T\times 3 \times H \times W}$ as input is fed into the encoder, where $T$ is temporal length and spatial resolution is $H\times W$. 
Following the protocol in ViT \citep{dosovitskiy2020vit} and TimeSformer \citep{bertasius2021timesformer}, the input video clip is divided into $T \times N$  non-overlapping spatio-temporal patches with spatial size of $P\times P$, where $N = HW/{P^2}$. 
In particular, the video clip $\mathbf{X}$ is fed through a 2D convolution layer to embed patch tokens $\mathbf{X}^z\in \mathbb{R}^{TN \times D}$.
%The patches $x\in \mathbb{R}^{T\times N \times 3 \times P \times P}$ are fed through a 2D convolutional layer to format tokens $z\in \mathbb{R}^{T N \times D}$.

To maintain the spatio-temporal information of video data at the same time during the training procedure, we append learnable temporal positional embeddings $\mathbf{e}^t\in \mathbb{R}^{T\times D}$ and spatial positional embeddings $\mathbf{e}^s\in \mathbb{R}^{N\times D}$ to each input token. 
All input patches within the $i^{th}$ ($i \in \lbrace0,1,...,T-1\rbrace $) frame (but different spatial locations) share the same temporal positional embedding $\mathbf{e}_i^t$, while all patches in the same $j^{th}$ spatial location ($j \in \lbrace0,1,...,N-1\rbrace $), regardless of their frame indexes, are given the same spatial positional embedding $\mathbf{e}_j^s$.
In our MAM$^2$, we adopt a ``random tube masking" strategy which simply broadcasts a 2-D random mask to all time steps of frames. 
%Since adjoining patches in space are coherent with a high masking ratio \citep{tong2022videomae}, space-only sampling may retain less information and produce an overly demanding pre-training challenge. 
Our MAM$^2$ takes mask ratio $\rho$ of $75\%$ to perform random tube masking. {It is illustrated in appendix.}

\textbf{Encoder.}
% 12 layers
The encoder is a stack of factorized space-time attention blocks \cite{bertasius2021timesformer} and it maps the embedded patches into their latent representations. 
The output of the space-time attention block is obtained by sequentially applying temporal and then spatial self-attention on its input.
This architecture can well learn spatio-temporal representation of video frames, in addition, it significantly reduces computation and memory complexity.

\textbf{Regressor and alignment.}
% 6 layers 
The process of regressor and latent representation alignment is analogous to context autoencoder~\citep{chen2022context}. 
Latent contextual regressor is a stack of cross-attention blocks. In particular, each cross-attention block takes the representations of visible patches as key and value and the mask queries as query to regress representations for masked patches. We make the mask queries learnable in the training procedure. 
%In this way, regressor predicts the latent representations on masked video patches so that decoders work over masked ($75\%$) latent representations with self-attention to reduce computation cost. 
Aligning the regressed representations of the masked patches with the groundtruth representations produced by the encoder is achieved with the following constraint: 
\begin{equation}\label{e1}
%\mathit{L_{Alignment}} = \frac{1}{M} \sum_{p\in M} | I(p) - \hat{I}(p) | ^{2},
%\mathcal{L}_{Alignment} = \frac{1}{|M|} \sum_{p\in M} | I_p - \hat{I}_p | ^{2},
\mathcal{L}_{\operatorname{Alignment}} = \frac{1}{|\mathcal{M}|} \sum_{p\in \mathcal{M}} \| \mathbf{r}_p - \hat{\mathbf{r}}_p \|_2^{2},
\end{equation}
where $\mathcal{M}$ is set of masked tokens, $|.|$ is the number of elements, $p$ is the masked token index, $\mathbf{r}_p$ is the regressed representation and $\hat{\mathbf{r}}_p$ is the alignment target which is produced by feeding the groundtruth masked patch into the encoder. 
Eventually, it is the regressed representations of the masked patches that are used for pre-text tasks completion. The regressor accompanied by alignment constraint ensures the encoder focuses on representative space-time feature abstraction and explicitly avoids the decoders involving in feature encoding. In this way, the encoder is driven by pretext tasks to excavate powerful spatial-temporal features.
%, the embedded tokens are flattened $z\in R^{T \times N \times D}$. 

%Through the use of cross-attention modules with visible tokens from the encoder, learnable mask queries are used to predict the latent representations on masked patches.
%The purpose of latent representation align module is to gradually guide the regressor to produce mask tokens from the teacher encoder that are similar to real ones. The teacher encoder shares the weights with the primary encoder during pre-training phase. To align the predicted mask tokens with the real mask tokens from the teacher encoder, we use MSE loss:

\textbf{Decoders.}
Two pretext tasks are specifically implemented by dual-stream decoders in our MAM$^2$. One is an appearance decoder to decode visual tokens for the masked patches, and the other is a motion decoder to predict the RGB difference for masked regions. 
%The details of target are presented in \ref{sec:targets}. 
Similar to the encoder, both decoders consist of a stack of factorized time and space attention blocks.
Latent representations based on masked queries generated by the regressor serve as the input for decoders, avoiding direct utilization of the information contained in the visible tokens.
As a result, the encoder is guided by the appearance and motion decoders to fully exploit spatio-temporal representation since it needs to simultaneously optimize the tasks of visual and motion prediction.   

\subsection{Pre-training Objective}

\paragraph{Appearance and Motion targets.}
\label{sec:targets}
For the appearance stream, we use a discrete variational autoencoder (dVAE) proposed in \citep{esser2021taming} to generate the discrete visual tokens of masked patches as the prediction targets of appearance decoder. On the other hand, we investigate three different types of motion targets on the motion stream:
\begin{itemize}
\item \textbf{Optical flow.} We extract on-the-fly optical flow by using a pre-trained deep network, Recurrent All-Pairs Field Transforms (RAFT) \citep{teed2020raft}, on the consecutive sampled frames. By using optical flow as a target, encoders can be forced to extract short-term motion cues between video frames more effectively.
However, using RAFT is computationally expensive, thus we investigate RGB difference as a more lightweight alternative.
% Optical flow as target forced encoder better to extract the short-term motion cues between video frames. However, introducing RAFT is computationally expensive and can not reveal a sufficiently attractive performance improvement compared to RGB difference.
\item \textbf{RGB difference.}
We simply compute the difference of the raw RGB values between adjacent video frames to obtain RGB difference target.
This is an effective and efficient way to capture short-term motion pattern and also provide local motion statistic as a complement to the appearance target. 
In particular, our method simply predict the RGB difference between $t$ th frame and $t-1$th frame on masked patches for $\forall t \in \lbrace1,2,...,T-1\rbrace $.

\item \textbf{Clip order prediction.}
In contrast to the motion modality discussed above, which focus on short-term motion cues, we explore long-range temporal prediction target. 
Specifically, we utilize the video clip order prediction task \citep{xu2019self, wang2021self} to predict the correct order of the randomly shuffled latent representations. The input of the motion decoder at each masked tube position is a tuple of temporally shuffled feature clips generated by the regressor for the masked tube, and we use a [CLS] token on motion decoder and its corresponding output representation is projected to a probability distribution over all possible feature clip orders. 
Actually, we divide and randomly shuffle the latent feature produced by the regressor to 2 clips (2 possible clip orders) at each masked tube.
\end{itemize}

\paragraph{Loss Function.}
For the appearance decoder, we use the off-the-shelf dVAE tokenizer to generate the discrete tokens as the targets. The discrete token label is assigned to each masked patch. 
Consequently, the target of appearance decoder is a $K$-class label where $K$ is 16384. After a linear prediction head $\mathbf{W}_{m}\in \mathbb{R}^{D\times K}$, where $D$ is 768 for ViT-Base, a softmax cross entropy loss is performed on the logits:
\begin{equation}\label{e2}
%\mathit{L_{Appearance}} = -\frac{1}{|M|}{\sum_{i\in M} {\sum_{j=1}^{K}} Y_{m(i,j)} log(P_{i,j})}
\mathcal{L}_{\operatorname{Appearance}} = -\frac{1}{|\mathcal{M}|} \sum_{i\in \mathcal{M}}{\sum_{j=1}^{K} {y_{m(i,j)} \log(p_{i,j})}},
\end{equation}
where $y_{m(i,j)}\in\{0, 1\}$ indicates the ground-truth label of sample $i$ is class $j$ or not, $p_{i,j}$ indicates predicted probability of sample $i$ in class $j$.

For motion decoder, we simply compute the RGB difference $\hat{\mathbf{D}}\in \mathbb{R}^{(T-1) \times \rho N \times C}$ of input $T$ frames at the masked regions. The target is to reconstruct the RGB difference of masked tokens. MSE loss is also adopted here to optimize this decoder:
\begin{equation}\label{e3}
%\mathit{L_{Motion}} = \frac{1}{M} \sum_{p\in M} | D(p) - \hat{D}(p) | ^{2}.
\mathcal{L}_{\operatorname{Motion}}  = \frac{1}{|\mathcal{M}'|} \sum_{p\in \mathcal{M}'} \| \mathbf{D}_p - \hat{\mathbf{D}}_p \|_2^{2}
\end{equation}
where $\mathcal{M}'$ is set of masked patches except for the ones of the $T^{th}$ frame, $|.|$ is the number of elements, p is the index of masked patches and $\mathbf{D}_p$ is the decoder prediction.
Finally, the hybrid loss is shown as follows:
\begin{equation}\label{e4}
\mathcal{L} = \mathcal{L}_{\operatorname{Appearance}} + \mathcal{L}_{\operatorname{Motion}} + \alpha \mathcal{L}_{\operatorname{Alignment}}
\end{equation}
where $\alpha$ is set to 2 in our experiments.
\section{Related work}
\textbf{Video Recognition.}
Research on video recognition has gone through rapid development, thanks to the availability of large-scale video datasets, e.g., Kinetics \citep{carreira2017k400,carreira2018k600,carreira2019k700} and Something-Something \citep{goyal2017something}. Convolutional networks have long dominated in video understanding tasks \citep{tran2015learning,tran2018closer,feichtenhofer2019slowfast,lin2019tsm,feichtenhofer2020x3d,wu2021mvfnet}. With the seminal work of Vision Transformer(ViT) \citep{dosovitskiy2020vit,touvron2021training}, researches on backbone architecture for video understanding have gradually shifited from CNNs to Transformers. TimeSformer \citep{bertasius2021timesformer} and ViViT \citep{arnab2021vivit} suggest the factorized spacetime attention for a strong speed-accuracy tradeoff. Video SwinTransformer \citep{liu2021videoswin} and MViT \citep{fan2021mvit} introduce the hierarchical structure to reduce computational cost. MTV \citep{yan2022multiview} separates encoders to represent each perspective of the input video. In this work, we follow the simple recipe of factorized spacetime attention and focus on exploring the pre-training of video transformer in a \textit{masked video modeling} manner.

\textbf{Self-supervised Video Representation Learning.}
For self-supervised video representation pre-training (SSVP), various pretext tasks are often designed based on the image contrastive self-supervised methods to learn spatiotemporal video representation including \citep{kuang2021video, yao2021seco, bai2020can, wang2021enhancing, fang2022mamico} and etc. Also, transformers remain the focus of video SSVP research, \citep{ranasinghe2022self} and \citep{wang2022long} explore intriguing properties of vision transformers \citep{dosovitskiy2020vit} in the video domain while \citep{recasens2021broaden, xu-etal-2021-vlm, xu-etal-2021-videoclip} jointly model multiple modalities with transformers. There are also some dense contrastive learning methods which have shown promissing results on video-related downstream tasks. In particular, DenseCL \citep{wang2021dense}, PixPro \citep{xie2021propagate} perform pairwise contrastive learning at pixel-level where corresponding pixels are encouraged to be consistent. 

On the other hand, masked visual autoencoders have also been proposed to learn effective video representations based on the mask-and-reconstruct pipeline due to its great success on image domain. Among the masked image modeling (MIM) pre-training schemes, MAE \citep{he2021mae}, BEiT \citep{bao2021beit} and CAE \citep{chen2022context} are most popular ones. 
There are concurrent works mostly related to ours.
ST-MAE \citep{feichtenhofer2022videomaefb} and VideoMAE \citep{tong2022videomae} directly extend MAE framework to the video domain by reconstructing the raw pixels of masked video patches from the visible contexts in a spacetime-agnostic manner. 
BEVT \citep{wang2022bevt} jointly reconstructs the discrete visual tokens obtained by an offline tokenizer in the image and video domain. 
In contrast to previous methods, our study firstly reveals that explicitly introducing motion cues related pretext task brings more efficient and transferable spatiotemporal representations learning. 
% To better realize our design of MAMM and make the encoder and decoder to perform their duties, the encoder-regressor-decoder in CAE\citep{chen2022context} framework is borrowed to our work.

\section{Experiments}
\subsection{Experimental Setups}

\textbf{Architecture.}
We begin with the vanilla ViT architectures \citep{dosovitskiy2020vit}, then we employ the factorzied space-time attention \citep{arnab2021vivit} to build our video encoder.
We adopt the ViT base and large architecture as our encoder.
% ViT-B (12 transformer layers with dimension 768) and ViT-L (24 transformer layers with dimension 1024). 
\emph{Please refer to Appendix for more details of the architectures.}
To minimize the parameter scale for positional embeddings, we add separable space and temporal positional embeddings to the patch tokens.

To jointly extract space-time features, we use a latent contextual regressor \citep{chen2022context} that consists of four transformer layers with cross-attention.
Two lightweight decoders are used to reconstruct visual tokens and the motion information of masked patches. 
Specifically, the appearance decoder is composed of four transformer layers based on factorised self-attention, just like the encoder, while the motion decoder is composed of two transformer layers. 
Additionally, the positional embeddings of the encoder are multiplexed onto the regressor and the decoders. 
%\emph{Please refer to Supplementary for more architectural details.}
%The specific architectural design is shown in Table \whwu{ref} .

\textbf{Implementation Details.}
In this paper, we evaluate our method on four widely used video datasets, \emph{i.e.}, Kinetics-400, Something-Something V2, UCF-101 and HMDB-51. \emph{The statistics of these datasets are provided in Appendix.}
We conduct all experiments on Something-SomethingV2 and Kinetics-400 datasets with 32 NVIDIA A100-40GB GPUs. As for the smaller UCF101 and HMDB51 datasets, we use 8 NVIDIA A100-40GB GPUs.
We follow BEVT \citep{wang2022bevt} to use the off-the-shelf visual tokenizer VQGAN \citep{esser2021taming} to generate semantic tokens, and employ the frame-level RGB difference as the ground-truth of motion prediction for masked video patches.
\emph{More implementation details are available in Appendix.}

\subsection{Ablation Studies}

We follow the paradigm of self-supervised pre-training and then fine-tuning in previous work \citep{feichtenhofer2022masked,tong2022videomae}, and conduct ablation studies to analyze the details of our design choices. Then, we investigate the effectiveness of different components of our framework.

%##################################################################################################

% overall table of all ablations
\begin{table}[t]
\makebox[\textwidth][c]{\begin{minipage}{1.1\linewidth}
\centering
%#################################################
% Decoder Depth
%#################################################
\subfloat[
\textbf{Depth of appearance decoder and motion decoder}. A shallower motion decoder compared with appearance decoder works better for a easier reconstruction target. 
\label{tab:depth_decoders}
]{
\begin{minipage}{0.3\linewidth}{\begin{center}
\tablestyle{3pt}{1.05}
\begin{tabular}{x{35}x{28}x{25}}
%\begin{tabular}{xxx}
appearance & motion & acc. \\
\shline
2 & 2 & 90.54  \\
4 & 4 & 90.97  \\
4 & 2 & \baseline{\textbf{91.09}}  \\
\end{tabular}
\end{center}}\end{minipage}
}
%\vspace{2em}
\hspace{1em}
%#################################################
% Sampling stride
%#################################################
\subfloat[
\textbf{Pretraining sampling stride}. The motion reconstruction target is more difficult for the larger sampling stride, and sampling stride of 4 works the best.
\label{tab:stride}
]{
\begin{minipage}{0.30\linewidth}{\begin{center}
\tablestyle{4pt}{1.05}
\begin{tabular}{x{40}x{45}}
stride & acc. \\
\shline
2 & 90.86 \\
4 & \baseline{\textbf{91.09}} \\ 
8 & 90.27 \\  
\end{tabular}
\end{center}}\end{minipage}
}
\hspace{1em}
%#################################################
% Regressor and Decoder Depth
%#################################################
\subfloat[
\textbf{Regressor and decoder width}. Matched regressor and decoder width with the encoder works the best in our framework.
\label{tab:decoder_width}
]{
\centering
\begin{minipage}{0.3\linewidth}{\begin{center}
\tablestyle{4pt}{1.05}
\begin{tabular}{x{40}x{40}}
dim &  acc. \\
\shline
384 & 90.08 \\
512 & 90.35 \\
768 & \baseline{\textbf{91.09}} \\
\end{tabular}
\end{center}}\end{minipage}
}
\hspace{1em}
\vspace{1em}
\\
%#################################################
% Mask strategy
%#################################################
\subfloat[
\textbf{Mask strategy}. 75\% mask ratio for tube masking works the best.
 \label{tab:mask}
]{
\begin{minipage}{0.3\linewidth}{\begin{center}
\tablestyle{3pt}{1.05}
\begin{tabular}{y{28}x{30}x{24}}
case & ratio & acc. \\
\shline
cube & 40\% & 89.87 \\
tube & 60\% & 90.26 \\
tube & 75\% & \baseline{\textbf{91.09}} \\
tube & 90\% & 87.76 \\
\end{tabular}
\end{center}}\end{minipage}
}
\hspace{1em}
%#################################################
% Tokenizer
%#################################################
\subfloat[
\textbf{Appearance reconstruction objective}. Results of using different tokenizers to form the appearance prediction targets.
 \label{tab:tokenizer}
]{
\begin{minipage}{0.3\linewidth}{\begin{center}
\tablestyle{3pt}{1.05}
\vspace{-7em}
\begin{tabular}{y{58}x{25}}
target & acc. \\
\shline
VQGAN-8192  & 90.88 \\
VQGAN-16384  & \baseline{\textbf{91.09}} \\
DALL-E & 90.61 \\
%pixel & 85.82 \\
\end{tabular}
\end{center}}\end{minipage}
}
\hspace{1em}
%#################################################
% Motion Target
%#################################################
\subfloat[
\textbf{Motion reconstruction objective}. Results of using different motion target as in Sec.~\ref{sec:targets}.
 \label{tab:motion}
]{
\begin{minipage}{0.3\linewidth}{\begin{center}
\tablestyle{3pt}{1.05}
\vspace{-7em}
\begin{tabular}{y{58}x{25}}
target & acc. \\
\shline
optical flow  & 91.05 \\
RGB-diff  & \baseline{\textbf{91.09}} \\
clip-order  & 90.12 \\
\end{tabular}
\end{center}}\end{minipage}
}
\hspace{1em}
\\
%#################################################
\caption{\textbf{Ablation experiments} of model design on UCF101. We uniformly pretrain 800 epochs and finetune 100 epochs. The encoder is ViT-B and the input video with the size of 16$\times$224$\times$224 is embeded to patch tokens with the size of 1$\times$16$\times$16. The default settings are marked in \colorbox{baselinecolor}{gray}.
\label{tab:ablations}
}
\vspace{-1.5em}
\end{minipage}}
\end{table}
%##################################################################################################

\textbf{Decoder Depth.}
\tblref{tab:depth_decoders} shows the performance of different  transformer block number in decoders. Intuitively, the more difficult reconstruction target needs a deeper decoder with better modeling capability and vice versa. The training objective for reconstructing RGB difference is easier than reconstructing the semantic tokens. The result also verifies that appearance decoder with 4 blocks and motion decoder with 2 blocks works better than appearance and motion decoder both with 4 blocks (91.09\% v.s. 90.97\%) or 2 blocks (91.09\% v.s. 90.54\%).

\textbf{Masking strategy.}
We start with the cube masking mechanism used by \citep{bao2021beit,tan2021vimpac,wang2022bevt} which a block-wise sampling mask is expanded along the temporal axis. Then, we study the temporal tube masking mechanism in VideoMAE ~\citep{tong2022videomae} where all frames share the same random spatial mask as MAE does~\citep{he2022masked}. 

Interestingly, we observe that random tube masking mechanism works better than block-wise masking mechanism (89.87\% \emph{v.s.} 91.09\%) in our framework. This is inconsistent with a series of token prediction methods in image and video domain, such as ~\citep{chen2022context}, ~\citep{bao2021beit} and ~\citep{wang2022bevt}. We also increase the masking ration from 60\% to 90\% based on tube masking mechanism and report the result in \tblref{tab:mask}. The tube masking with 75\% ratio works the best (91.09\%). A higher marking ratio such as 90\% performs poor (87.76\%). A possible reason is that visual token prediction is more difficult than pixel reconstruction and more context is needed for predicting masked discrete tokens, such that the task can be completed with meaningful supervision signals for the encoder. 

%#################################################################################################
\begin{table*}[t!]
\centering
\tablestyle{5.0pt}{1.04}
%\begin{tabular}{x{60}|x{65}|x{50}x{35}|x{30}x{28}}
\begin{tabular}{l|p{25mm}<{\centering}|p{14mm}<{\centering}p{15mm}<{\centering}|p{12mm}<{\centering}|p{12mm}<{\centering}|p{8mm}<{\centering}}
\shline
\multirow{2}*{case} & \multirow{2}*{1 decoder 2 tasks} & \multicolumn{2}{c}{decoupled decoders} \vline & \multirow{2}*{regressor}& \multirow{2}*{alignment} & \multirow{2}*{acc.} \\
\cline{3-4} 
{} & {} & semantic &motion & & \\
\shline
MAM$^2$-c         & \cmark &\xmark &\xmark &\cmark &\cmark & 89.82 \\ \hdashline%\hline
w/o appearance  & \xmark &\xmark &\cmark &\cmark &\cmark &87.56 \\
w/o motion    & \xmark &\cmark &\xmark &\cmark &\cmark &90.03 \\ \hdashline%\hline
w/o regressor & \xmark &\cmark &\cmark &\xmark &\cmark &90.84 \\
w/o alignment & \xmark &\cmark &\cmark &\cmark &\xmark &90.63 \\ \hdashline%\hline
MAM$^2$           & \xmark &\cmark &\cmark &\cmark &\cmark &\baseline{\textbf{91.09}} \\ \shline
\end{tabular}
\caption{\textbf{Empirical study on impact of each main components of our MAM$^2$ framework on UCF101 dataset.} The default settings are marked in \colorbox{baselinecolor}{gray}.}
\vspace{-8mm}
\label{tab:components}
\end{table*}
%#################################################################################################
\textbf{Temporal sampling stride.}
For the branch of our motion decoder, the temporal sampling stride of video frames reflects the prediction difficulty when making predictions for the masked video patches. We provide the results of different temporal sampling stride in \tblref{tab:stride}. Stride of 2 and 8 both obtain slightly worse results (90.86\% and 90.27\%). Stride of 4 achieves the trade-off between the pretext task difficulty and the fine-tuning accuracy (91.09\%).

\textbf{Regresor and Decoder width.}
\tblref{tab:decoder_width} report the result of different regresor and docoders depth. Note that, the depth of regressor, semantic decoder and the modiotn decoder are the same. In particular, we decrease the width from 512 to 384. We add a projection layer after the encoder in order to project the latent representation to the width of regressor for the case of 512 or 384. And the width of 768 achieve the best results of 91.09\%.

% To appendix???
\textbf{Appearance prediction objective.}
%\textcolor{red}{pixel result is counterintuitive, delete?}
We ablate different prediction targets for the appearance decoder and show results in \tblref{tab:tokenizer}. We use the visual tokenizer of VQGAN \citep{esser2021taming} that pre-trained on OpenImages with GumbelQuantization. The VQGAN tokenizer with a vocabulary size of 16384 achieves the best performance (91.09\%). Using the VQGAN-8192 tokenizer results in a slight drop of 0.21\%. A discrete variational autoencoder (dVAE) in \citep{ramesh2021zero} performs much worse than VQGAN (90.61\%).
Please refer to appendix for tokenizer details.

\textbf{Motion prediction objective.}                   
We further conduct studies to evaluate different motion targets as introduced in Sec.~\ref{sec:targets} and results are summarized in Table \ref{tab:motion}. The optical flow achieves the similar performance with RGB difference, however it introduces extra computational cost due to RAFT. Compared to optical flow, RGB-diff is a practical and efficient target for capturing the short-term motion cues. On the other hand, clip-order prediction (90.12\%) providing long-term temporal evolution cues is unproductive compared to using only the appearance stream (90.03\%, see ``w/o motion'' in Table \ref{tab:components}). We attribute it to the position embedding added into the masked queries when querying the latent representations for masked patches by the regressor, which will collapse the clip-order prediction task, the position information implied in the latent representations can lead to leakage of original temporal information.

\textbf{Contributions of the main components in MAM$^2$.}
In this subsection, we explore the contributions of the three main components in MAM$^2$ architecture, \textit{i.e.}, the regressor, the disentangled decoders and the alignment constraint, and report the results on UCF101 dataset in \tblref{tab:components}. Firstly, we compare the performance of coupled decoder (\textit{i.e.}, case MAM$^2$-c where only a single decoder with 2 prediction heads is used for the 2 pretext tasks) and disentangled decoders (case MAM$^2$). 
%In particular, we use a shared coupling decoder for visual token reconstruction and motion reconstruction. And we add two parallel linear heads followed by the coupling decoder to perform the corresponding reconstruction tasks. 
Results show that coupled decoder decreases the accuracy to 89.82\% compared to default settings (91.09\%).

Following the setting of disentangled decoders, we separately remove the the components of appearance decoder, motion decoder, regressor and the alignment constraint. It is clear that decoupling the appearance and motion decoders helps to improvement the overall performance significantly. In particular, only using the appearance decoder (w/o motion) result in a drop of 1.06\% compared to default settings, and only using the motion decoder (w/o appearance) drops significantly (3.53\%). Moreover, the w/o alignment achieves the result of \textcolor{black}{90.63\%} by removing the alignment constraint. The w/o regressor achieves the result of \textcolor{black}{90.84\%} by removing latent contextual regressor. These results imply that regressor together with the alignment loss is important to make the encoder focus on spatial-temporal feature excavation.

\begin{table*}[t!]
\centering
% \footnotesize
\tablestyle{2.0pt}{1.}
\begin{tabular}{l|c|c|c|c|c|c|c}
\shline
\textbf{Method} & \textbf{Backbone} & \textbf{Pre-train data} & \textbf{Frames} & \textbf{Views}  & \textbf{Param} & \textbf{Top-1}  & \textbf{Top-5} \\
\shline\hline
\multicolumn{1}{l}{\textsl{Supervised pre-training} } \\ \hdashline
NL I3D~\citep{wang2018nonlocal} & ResNet101 & IN-1K     & 128 & 10\x3  & 62  & 77.3 & 93.3    \\ %\multirow{4}{*}{ImageNet-1K}
TAM~\citep{liu2021tam} &  ResNet152 &IN-1K  &16 & 4\x3 & 59 & 79.3 & 94.1 \\
TDN$_{En}$~\citep{wang2021tdn} & ResNet101$_{\times 2}$ &IN-1K    & 8+16 & 10\x3 & 88 & 79.4 & 94.4   \\
Video Swin~\citep{liu2021videoswin}& Swin-B &IN-1K & 32 & 4\x3 & 88 & 80.6 & 94.6  \\
%\hline
TimeSformer~\citep{bertasius2021timesformer} & ViT-B &IN-21K   & 8 & 1\x3 & 121 & 78.3 &  93.7 \\ %\multirow{6}{*}{ImageNet-21K}
TimeSformer~\citep{bertasius2021timesformer} & ViT-L &IN-21K     & 96 & 1\x3 & 430 & 80.7 & 94.7 \\
ViViT FE~\citep{arnab2021vivit} & ViT-L &IN21K     & 128 & 1\x3 & N/A & 81.7 & 93.8 \\
Motionformer~\citep{patrick2021motionformer} & ViT-B &IN-21K       & 16 & 10\x3 & 109  & 79.7 & 94.2 \\
Motionformer~\citep{patrick2021motionformer} & ViT-L &IN-21K       & 32 & 10\x3 & 382  & 80.2 & 94.8 \\
Video Swin~\citep{liu2021videoswin}  & Swin-L &IN-21K      & 32 & 4\x3 & 197 & 83.1 & 95.9  \\
%\hline
ViViT FE~\citep{arnab2021vivit} & ViT-L & JFT-300M     & 128 & 1\x3  & N/A & 83.5 & 94.3 \\
ViViT~\citep{arnab2021vivit} & ViT-H & JFT-300M    & 32 & 4\x3 & N/A & 84.9 & 95.8 \\
%\hline
ip-CSN~\citep{tran2019csn} & ResNet152 &K400     & 32 & 10$\times$3 & 33 & 77.8 & 92.8 \\ %\multirow{4}{*}{\emph{no external data}}
SlowFast~\citep{feichtenhofer2019slowfast} &  R101+NL &K400      & 16+64 & 10\x3 & 60 & 79.8 & 93.9 \\
MViTv1~\citep{fan2021mvit} & MViTv1-B &K400     & 32 & 5\x1 & 37  & 80.2 & 94.4 \\
\hline
\multicolumn{1}{l}{\textsl{Self-supervised pre-training} } \\ \hdashline
VIMPAC~\citep{tan2021vimpac} & ViT-L & \scriptsize{HowTo100M+DALLE}    & 10 & 10\x3 & 307 & 77.4 & N/A \\
BEVT~\citep{wang2022bevt}  & Swin-B & \scriptsize{IN-1K+K400+DALLE}    & 32 & 4\x3 & 88 & 80.6 & N/A  \\
MaskFeat~\citep{wei2022masked} & MViT-L &K400     & 16  & 10\x1 & 218 & 84.3 & 96.3 \\
%\textcolor{gray}{MaskFeat{\scriptsize\textuparrow352}~\citep{wei2022masked}} & \textcolor{gray}{MViT-L} &  \textcolor{gray}{K600}  & \textcolor{gray}{40} & \textcolor{gray}{3790\x4\x3} & \textcolor{gray}{218} & \textcolor{gray}{{87.0}} & \textcolor{gray}{{97.4}} \\
VideoMAE~\citep{tong2022videomae} & ViT-B & K400 & 16 & 5\x3 & 87  & 80.9 & 94.7 \\
VideoMAE~\citep{tong2022videomae} & ViT-L & K400 & 16 & 5\x3 & 305 & 84.7 & 96.5 \\
OmniMAE~\citep{girdhar2022omnimae} & ViT-B & IN-1K+K400 & 16 & 5\x3 & 87  & 80.6 & N/A  \\
OmniMAE~\citep{girdhar2022omnimae} & ViT-L & IN-1K+K400 & 16 & 5\x3 & 305 & 84.0 & N/A  \\
ST-MAE\citep{feichtenhofer2022masked} & ViT-B & K400 & 16 & 7\x3 & 87  & 81.3 & 94.9 \\
ST-MAE\citep{feichtenhofer2022masked} & ViT-L & K400 & 16 & 7\x3 & 304 & 84.8 & 96.2 \\
%\textcolor{gray}{MAE\citep{feichtenhofer2022masked}} & \textcolor{gray}{ViT-L} &  \textcolor{gray}{K600}  & \textcolor{gray}{16} & \textcolor{gray}{598\x7\x3} & \textcolor{gray}{304} & \textcolor{gray}{{86.5}} & \textcolor{gray}{{97.2}} \\
\shline%\hline
\textbf{MAM$^2$ (Ours)} & ViT-B & K400+VQGAN & 16 & 7\x3 & 114 & \textbf{82.3} & \textbf{95.3} \\
\textbf{MAM$^2$ (Ours, 600epoch)} & ViT-L & K400+VQGAN & 16 & 7\x3 & 403 & \textbf{84.8} & \textbf{96.4} \\
\textbf{MAM$^2$ (Ours, 800epoch)} & ViT-L & K400+VQGAN & 16 & 7\x3 & 403 & \textbf{85.3} & \textbf{96.7} \\ \shline

\end{tabular}
\vspace{-0.5mm}
\caption{\textbf{Comparison with the state-of-the-art methods on Kinetics-400}. MAM$^2$ is pre-trained 800 epochs for ViT-B and ViT-L. ``N/A" indicates the numbers are not available for us.}
%\vspace{-6pt}
\label{tab:k400}
\end{table*}
%#################################################################################################
%#################################################################################################
\begin{table*}[t]
\centering
% \scalebox{1.0}{
\tablestyle{2.0pt}{1.04}
\begin{tabular}{l|c|c|c|c|c|c|c}
\shline
\textbf{Method} & \textbf{Backbone} & \textbf{Pre-train data} & \textbf{Frames} & \textbf{Views}  & \textbf{Param} & \textbf{Top-1}  & \textbf{Top-5} \\
\shline\hline
\multicolumn{1}{l}{\textsl{Supervised pre-training} } \\ \hdashline
TSM$_{En}$~\citep{lin2019tsm}  & \footnotesize{ResNet50$_{\times 2}$}  & {IN-1K}   & 16+16 & 2\x3 & 49 & 66.0 & 90.5 \\  %\multirow{3}{*}{ImageNet-1K}
TAM~\citep{liu2021tam} &  \footnotesize{ResNet50$_{\times 2}$} &{IN-1K}    & 8+16 & 2\x3 & 51 & 66.0 & 90.1 \\
TDN$_{En}$~\citep{wang2021tdn} & \footnotesize{ResNet101$_{\times 2}$} & {IN-1K}     & 8+16 & 1\x3 & 88 & 69.6 & 92.2   \\

SlowFast~\citep{feichtenhofer2019slowfast} &  ResNet101 & {Kinetics-400} & 8+32 & 1\x3 & 53 & 63.1 & 87.6 \\ %\multirow{2}{*}{Kinetics-400}
MViTv1~\citep{fan2021mvit} & MViTv1-B &{Kinetics-400} & 64 & 1\x3 & 37 & 67.7 & 90.9  \\

TimeSformer~\citep{bertasius2021timesformer} & ViT-B & {IN-21K}  & 8 & 1\x3 & 121 & 59.5 &  N/A \\ %\multirow{2}{*}{ImageNet-21K} 
TimeSformer~\citep{bertasius2021timesformer} & ViT-L &{IN-21K} & 64  & 1\x3 & 430 & 62.4 & N/A \\

ViViT FE~\citep{arnab2021vivit} & ViT-L &  IN-21K+K400  & 32 & 4\x3 & N/A & 65.9 & 89.9 \\ %\multirow{4}{*}{\footnotesize IN-21K+K400}
Motionformer~\citep{patrick2021motionformer} & ViT-B &  IN-21K+K400    & 16 & 1\x3 & 109  & 66.5 & 90.1 \\
Motionformer~\citep{patrick2021motionformer} & ViT-L & IN-21K+K400   & 32 & 1\x3 & 382  & 68.1 & 91.2 \\
Video Swin~\citep{liu2021videoswin}  & Swin-B & IN-21K+K400  & 32 & 1\x3 & 88 & 69.6 & 92.7  \\
\hline
\multicolumn{1}{l}{\textsl{Self-supervised pre-training} } \\ \hdashline

VIMPAC~\citep{tan2021vimpac} & ViT-L & HowTo100M &10 & 10\x3 & 307 & 68.1 & N/A \\
BEVT~\citep{wang2022bevt}  & Swin-B & \scriptsize{IN-1K+K400+DALL-E}  & 32 & 1\x3 & 88 & 70.6 & N/A  \\ %\scriptsize{IN-1K+K400+DALLE} 
OmniMAE\citep{girdhar2022omnimae} & ViT-B & IN-1K+SSv2 &16  & 2\x3 & 87 & 69.5 &N/A \\
%    
%\textcolor{gray}{MaskFeat{\scriptsize\textuparrow312}~\citep{wei2022masked}} & \textcolor{gray}{MViT-L} & \textcolor{gray}{Kinetics-600} & \textcolor{gray}{40}  & \textcolor{gray}{2828\x1\x3} & \textcolor{gray}{218} & \textcolor{gray}{75.0} & \textcolor{gray}{95.0} \\
%VideoMAE & ViT-B & \multirow{3}{*}{\emph{no external data}} & 16 & 180\x2\x3 & 87 & 70.6 & 92.7 \\
VideoMAE~\citep{tong2022videomae} & ViT-B & SSv2 & 16 & 2\x3 & 87 & 70.6 & 92.7 \\
\shline%\hline
\textbf{MAM$^2$ (Ours)} & ViT-B & SSv2+VQGAN & 16 &2\x3 & 114 &\textbf{71.3}  &\textbf{93.1} \\ \shline

\end{tabular}
\caption{\textbf{Comparison with the state-of-the-art methods on Something-Something V2.} MAM$^2$ is pre-trained 1200 epochs for ViT-B. ``N/A" indicates the numbers are not available for us.}
\label{tab:ssv2}
\vspace{-12pt}
\end{table*}
%#################################################################################################

%#################################################################################################
\begin{figure*}[h]
  \centering
  \includegraphics[width=0.9\textwidth]{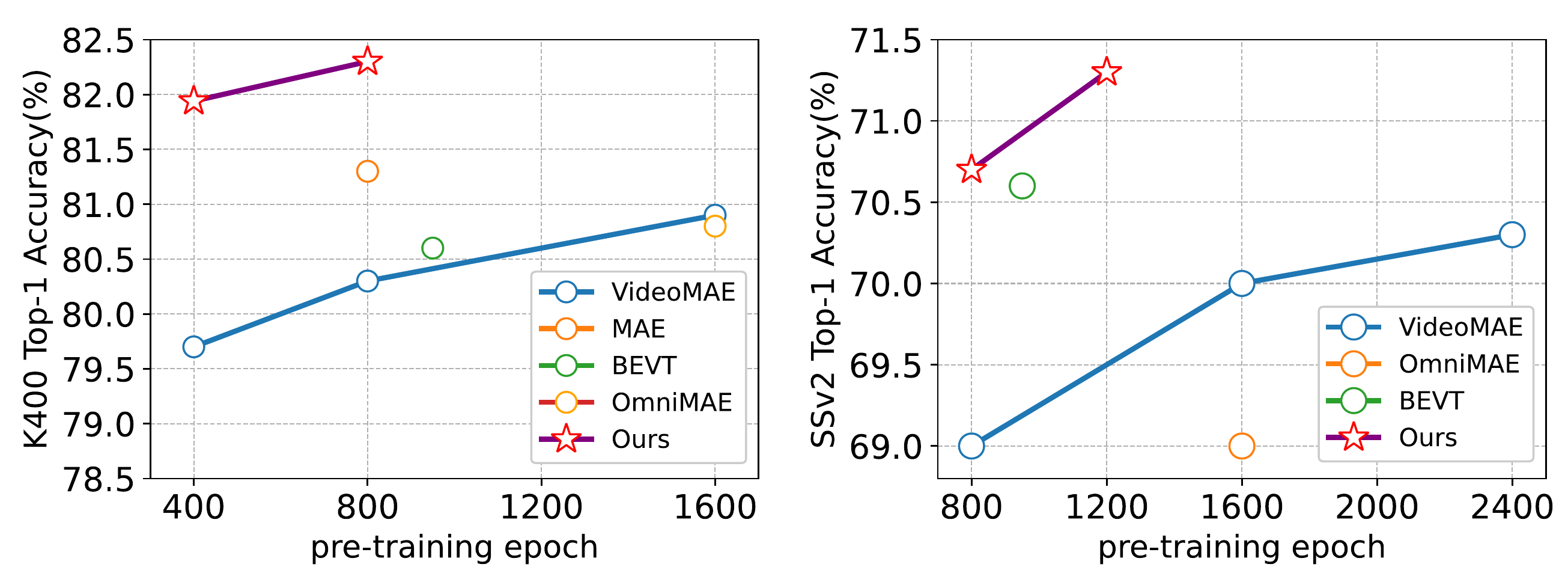}
  \caption{Top-1 accuracy on Kinetics-400 and Something-Something V2 with different masked video modeling methods. The default architecture is ViT-B. (\textit{Best viewed in color.})}
  \label{fig:compare}
\end{figure*}
%#################################################################################################

\subsection{Main Results and Analysis}

\textbf{Comparisons with State-of-the-art methods.}
We report the result on Kinetics-400 and Something-Something V2 datasets in \tblref{tab:k400} and \tblref{tab:ssv2}. Compare with the previous state-of-the-art methods, firstly, our MAM$^2$ achieves a gain of around +1.0\% compared with the SOTA results of ST-MAE on Kinetics-400 with the ViT-B architecture. On the temporal-heavy dataset of Something-Something V2, MAM$^2$ (ViT-B) can obtain a better performance of 71.3\% compared to the VideoMAE (70.6\%). Secondly, MAM$^2$ respectively achieves the pre-training with 800 and 1200 epochs on Kinetics-400 and Something-Something V2 that is 2$\times$ fewer than the VideoMAE, as shown in Fig.\ref{fig:compare}. Thirdly, we report the results on two other small datasets (UCF101 and HMDB51), and compared with the previous methods. \tblref{tab:ucf_hmdb} shows that MAM$^2$ outperforms VideoMAE by 0.7\% with 2.7$\times$ fewer pre-training epochs (1200 v.s. 3200), and obatain the gain of 1.4\% with 3$\times$ fewer pre-training epochs (1600 v.s. 4800). At last, we observe that our MAM$^2$ with ViT-L also outperforms ST-MAE and VideoMAE when they are all pre-trained for 800 epochs, we can also achieve comparable results with ST-MAE or VideoMAE with only 600 epochs.

\iffalse
%Appendix ？
%#################################################################################################
\begin{wraptable}[5]{r}{0.45\textwidth}
\setlength{\tabcolsep}{0.2em}
\centering
\vspace{-5pt}
\resizebox{0.45\textwidth}{!}{%
\begin{tabular}{l|cccc}
\toprule
{method} & {epoch} & {from scratch} & {lin.acc.}  & {ft.acc.} \\
\hline
MoCo v3     &300       &32.6        &33.7     &54.2 \\
VideoMAE    &800       &32.6        &38.9     &69.3 \\
MAM$^2$    &400       &  --        &--       &70.4 \\
\bottomrule
\end{tabular}%
}
\vspace{-0.5mm}
\caption{ xxx
}
%\vspace{-10pt}
\label{tab:ss_compare}
\end{wraptable}
%#################################################################################################

Benefit from the clever design of decoupling decoders, our method efficiently process spatio-temporal redundant information. We compare to previous self-supvised pre-training methods on Something-Something V2. Our method surpasses VideoMAE by +1.1\% with 2$\times$  fewer pre-training epochs (as presented in \tblref{tab:ss_compare}) and achieve the gain of 15\% compared to the MoCo v3 \citep{chen2021empirical}. 
\fi

%#################################################################################################
\begin{table*}[t!]
\centering
\tablestyle{2.0pt}{1.04}
\begin{tabular}{l|c|c|c|c}
\shline
\textbf{Method} & \textbf{Backbone} & \textbf{Pre-train data} & \textbf{UCF101}  & \textbf{HMDB51} \\
\shline\hline

CoCLR~\citep{coclr} & S3D-G & UCF101 & 81.4 & 52.1 \\
Vi$^2$CLR~\citep{diba2021vi2clr} & S3D & UCF101 & 82.8 & 52.9 \\
MCN ~\citep{lin2021self} & R3D &UCF101/HMDB51 & 85.4 &54.8 \\
VideoMAE~\citep{tong2022videomae}  & ViT-B &UCF101/HMDB51 & 90.8 & 61.1\\
\hline
\textbf{MAM$^2$ (Ours)} & ViT-B & UCF101/HMDB51 &\textbf{91.5}  &\textbf{62.5} \\ \shline

\end{tabular}
\caption{\textbf{Comparisons to Previous Methods on UCF101 and HMDB51.} MAM$^2$ is pre-trained 1200 epochs on UCF101 and 1600 epochs on HMDB51.}
\vspace{-12pt}
\label{tab:ucf_hmdb}
\end{table*}
%#################################################################################################
\section{Limitation}
Our Masked Appearance-Motion Modeling (MAM$^2$) framework excites the encoder efficiently exploring spatiotemporal information in video data and significantly improved performance on downstream video recognition tasks. In this work, subjected to GPU resources, we do not explore how our MAM$^2$ would behave under the circumstances of larger pre-training batch size or larger pre-training epochs. 
%MAMM is conducted with 32 A100 GPUs which is less than 2$\times$ than VideoMAE (64 V100 GPUs) and ST-MAE (128 A100 GPUs) during pre-training and finetuning phases. This limits us to pre-train and fine-tune with larger batch size and larger backbones. 
On the other hand, our pre-training framework introduces an extra tokenizer, which also impacts the overall performance a lot, besides, the on-the-fly computation of the tokenizer further limits our pre-training batch size. The recent work \citep{bai2022masked} inspires us to revisit the appearance stream in a masked representation modeling manner to deprecate tokenizer meanwhile providing reciprocal information with motion targets, we leave this as our future work.

\section{Conclusion}
In this paper, we make the first attempt to investigate motion cues explicitly in the Masked Video Modeling (MVM) pre-training scheme. With our MAM$^2$ framework, disentangled appearance and motion decoders are dedicated to reconstruct visual and motion knowledge respectively, meanwhile the encoder is forced to learn robust and generalized spatiotemporal video representations. The disentangled decoders help to reduce the convergence of pre-training. Experimental results show that our method achieves competitive results on all video action recognition datasets with even a half number of pre-training epochs of state-of-the-art MVM methods.



\bibliography{iclr2023_conference}
\bibliographystyle{iclr2023_conference}

\clearpage
\appendix
\section{Datasets}
We evaluate our MAM$^2$ on four common video recognition datasets: Kinetics-400 \citep{carreira2017k400}, Something-Something V2 \citep{goyal2017something}, UCF101 \citep{soomro2012ucf101} and HMDB51 \citep{kuehne2011hmdb}. In Table \tblref{tab:datasets}, we report the key statistics of these four datasets. We demonstrate the effectiveness of our MAM$^2$ on both Spatially-heavy datasets such as Kinetics-400, UCF101 and HMDB51, and Temporally-heavy dataset, e.g. Something-Something V2.

%#################################################################################################
\begin{table*}[t!]
\small
\centering
\tablestyle{2.0pt}{1.04}
\begin{tabular}{l|c|c|c|c}
\shline
%\hline
{} & \textbf{Kinetics-400} & \textbf{Something-Something V2} & \textbf{UCF101}  & \textbf{HMDB51} \\
\shline
Training Split &240k &169k &9.5k &3.5k \\
Validation Split &20k &20k &3.5k &1.5k \\
Number of Classes &400 &174 &101 &51 \\
Average Video Duration &10s &4s &7s &4s \\

\shline
\end{tabular}
\caption{\textbf{The statistics for our used video recognition datasets.}}
%\vspace{-2em}
\label{tab:datasets}
\end{table*}
%#################################################################################################

\section{Implementation Details}
In the pre-training phase, we consistently use the dense sampling strategy in \citep{feichtenhofer2019slowfast} with stride 4. Moreover, we perform linearly scaling strategy for the base learning rate w.r.t $lr = base\ learning\  rate \times batch\  size / 256$.
During the pre-training phase, we adopt the repeated augmentation trick \citep{hoffer2020augment} for a fair comparison with \citep{tong2022videomae, feichtenhofer2022videomaefb}.
We uniformly sample 16 frames during the pre-training and finetuning phases for all datasets.
Moreover, we use frames extracted offline from the raw videos during pre-traing phase to reduce the data loading time, and we load the raw videos on the fly during fine-tuning phase.

\textbf{Kinetics-400.} Our settings follow \citep{feichtenhofer2022videomaefb,tong2022videomae}. We trained MAM$^2$ for 800 epochs in the pre-training phase. During the fine-tuning phase, we use the dense sampling with stride 4. For a fair comparison, we perform the inference protocol with 7 clips x 3crops following \citep{feichtenhofer2022videomaefb}. \tblref{tab:pretraining} shows the pre-training settings for the architecture ViT/B and ViT/L and \tblref{tab:finetune} shows the fine-tuning details.

\textbf{Something-Something V2.}
We trained MAM$^2$ based on ViT-B for 1200 epochs in the pre-training by default. During the fine-tuning phase, we perform the uniform sampling strategy with stride 2 following TSN \citep{wang2018tsn} and we adopt the inference protocol with 2 clips x 3crops for comparison with \citep{tong2022videomae}. Default settings for ViT-B in pre-training and fine-tuning are correspondingly shown in \tblref{tab:pretraining} and \tblref{tab:finetune}. Notably, $flip\  augmentation$ is deprecated during both the pre-training and fine-tuning phase.

\textbf{UCF101.}
Our MAM$^2$ is pre-trained for 800 epochs by default in ablation study and we report the final result with 1200 epochs in \tblref{tab:ucf_hmdb}. The $base\  learning\  rate$ is set to $1.5e\text{-}4$ during the pre-traing and is set to $1e\text{-}4$ during the fine-tuning. The batch size in single node is set to 32 for pre-training and fine-tuning. We adopt dense sampling with temporal stride of 4 for pre-traing and fine-tuning. Default settings in pre-training and fine-tuning are correspondingly shown in \tblref{tab:pretraining} and \tblref{tab:finetune}.

\textbf{HMDB51.}
Our MAM$^2$ is pre-trained for 1600 epochs by default and we report the final result in Table \tblref{tab:ucf_hmdb}. The $base\  learning\  rate$ is set to $1.5e\text{-}4$ during the pre-traing and is set to $1e\text{-}4$ during the fine-tuning. The batch size in single node is set to 32 for pre-training and fine-tuning. We adopt dense sampling with temporal stride of 4 for pre-training and fine-tuning.
Default settings for ViT-B in pre-training and fine-tuning are correspondingly shown in \tblref{tab:pretraining} and \tblref{tab:finetune}.

%##################################################################################################
\begin{table}[t!]
\tablestyle{1pt}{1.02}
%\scriptsize
%\footnotesize
\small
\begin{tabular}{y{80}|x{58}x{58}x{58}x{58}}
config & Kinetics-400 & Sth-Sth V2 & UCF101 & HMDB51 \\
\shline
optimizer & \multicolumn{4}{c}{AdamW \citep{loshchilov2017decoupled}} \\ 
base learning rate & \multicolumn{4}{c}{1.5e-4}\\
weight decay & \multicolumn{4}{c}{0.05} \\
optimizer momentum & \multicolumn{4}{c}{$\beta_1, \beta_2{=}0.9, 0.95$ \citep{chen2020generative}} \\
%batch size & \multicolumn{2}{c}{128} \vline & \multicolumn{2}{c}{32} \\
batch size &{128} &{128} &{32} &{32} \\
learning rate schedule & \multicolumn{4}{c}{cosine decay \citep{loshchilov2016sgdr}} \\
pre-training epochs & 800 & 1200 & 1200 & 1600 \\
warmup epochs &{20} &{30} &{30} &{40} \\
flip augmentation & \emph{yes} & \emph{no} & \emph{yes} & \emph{yes} \\
augmentation & \multicolumn{4}{c}{MultiScaleCrop} \\
\end{tabular}
%\vspace{.5em}
\caption{\textbf{Pre-training setting.}}
\label{tab:pretraining} %\vspace{-.5em}
%\vspace{-2mm}
\end{table}
%##################################################################################################

%##################################################################################################
\begin{table}[t!]
\tablestyle{1pt}{1.02}
%\scriptsize
%\footnotesize
\small
%\normalsize
\begin{tabular}{y{80}|x{58}x{58}x{58}x{58}}
config & Kinetics-400 & Sth-Sth V2 & UCF101 & HMDB51 \\
\shline
optimizer & \multicolumn{4}{c}{AdamW \citep{loshchilov2017decoupled}} \\
base learning rate & 5e-4(B) 1e-3(L) & 5e-4 & 1e-4 & 1e-4 \\
weight decay & \multicolumn{4}{c}{0.05} \\
optimizer momentum & \multicolumn{4}{c}{$\beta_1, \beta_2{=}0.9, 0.999$ \citep{chen2020generative}} \\
layer-wise lr decay  & \multicolumn{4}{c}{0.75 \citep{chen2020generative}} \\
%batch size & \multicolumn{2}{c}{512(B), 256(L)} \\
batch size & 256(B) 64(L) & 256 & 24 & 24 \\
learning rate schedule & \multicolumn{4}{c}{cosine decay \citep{loshchilov2016sgdr}} \\
warmup epochs & \multicolumn{4}{c}{5} \\
training epochs & 100(B) 35(L) & 35 & 100 & 60 \\
flip augmentation & \emph{yes} & \emph{no} & \emph{yes} & \emph{yes} \\
RandAug  & \multicolumn{4}{c}{(9, 0.5) \citep{cubuk2020randaugment}} \\
label smoothing  & \multicolumn{4}{c}{0.1 \citep{szegedy2016labelsmooth}}  \\
mixup  & \multicolumn{4}{c}{0.8 \citep{zhang2017mixup}} \\
cutmix  & \multicolumn{4}{c}{1.0 \citep{yun2019cutmix}} \\
drop path  & \multicolumn{4}{c}{0.1 \citep{huang2016droppath}} \\
repeated sampling &\multicolumn{4}{c}{2 \citep{hoffer2020augment}} \\
\end{tabular}
%\vspace{.5em}
\caption{\textbf{Fine-tuning setting.}}
\label{tab:finetune} %\vspace{-2.5mm}
\end{table}
%##################################################################################################

\section{Architecture Details}
We complement more architecture details of MAM$^2$ following Sec.~\ref{sec:architecture}. We design an encoder-regressor-decoder architecture with dual stream decoders for self-supervised video transformer pre-training. \tblref{tab:architecture} takes the 16-frame ViT-Base for example and displays the pipeline details during the pre-training phase.

%#################################################################################################
\begin{figure*}[h]
  \centering
  \includegraphics[width=\textwidth]{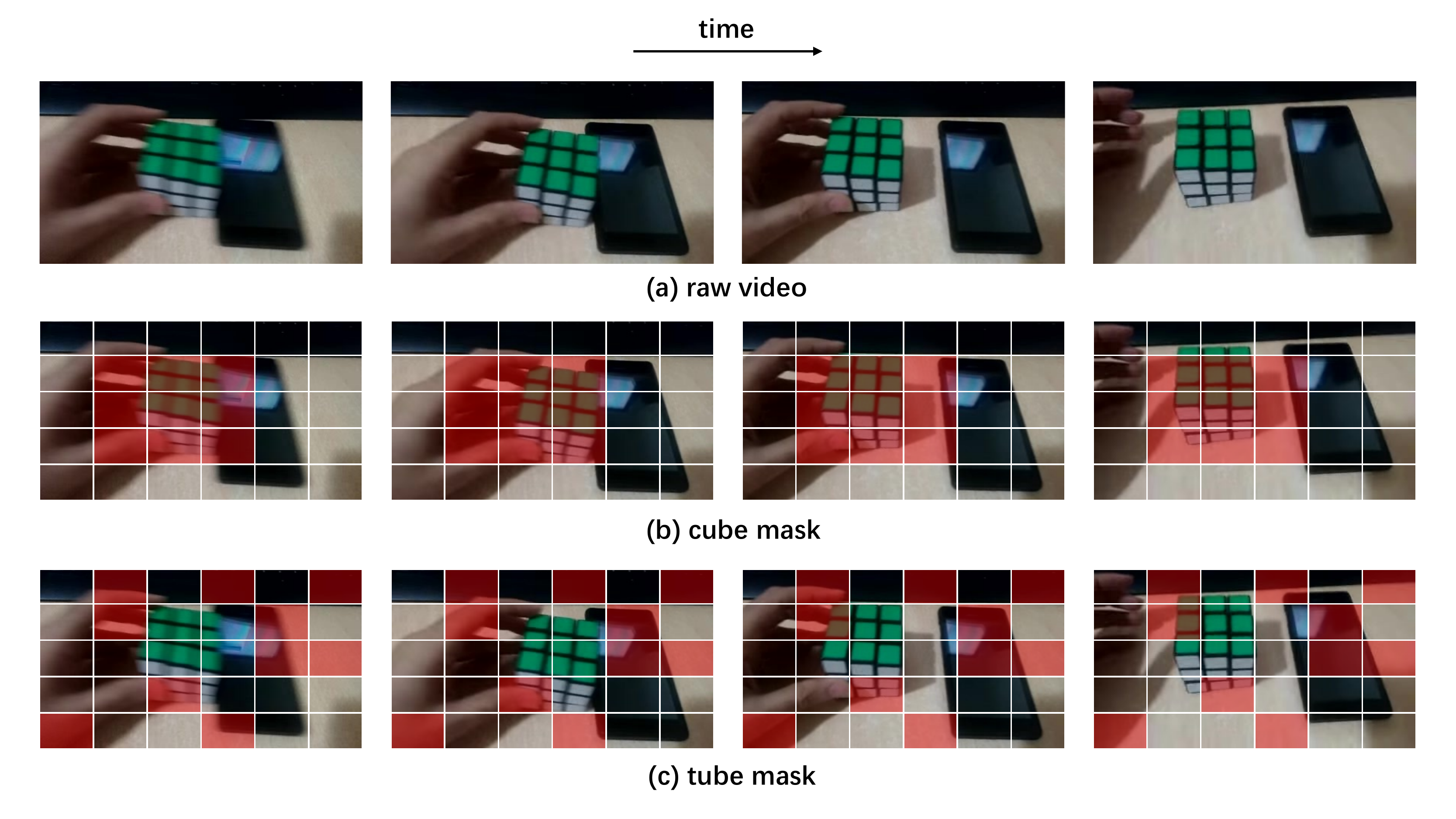}
  \caption{Demonstration of different masking strategies. (a) is the raw input video. (b) is \emph{cube} masking strategy and  (c) is \emph{tube} masking strategy. The masked patches are marked in red.}
  \label{fig:mask}
\end{figure*}
%#################################################################################################

%##################################################################################################
\begin{table}[t!]
\tablestyle{1pt}{1.08}
%\scriptsize
%\footnotesize
\small
\begin{tabular}{y{80}|x{110}|x{100}}
%\begin{tabular}{c|c|c}

\textbf{stage} & \textbf{component} & \textbf{output sizes} \\
\shline
data & \emph{temporal stride} 4 &{3}\x{16}\x{224}\x{224}  \\ \hline
patch embedding & \emph{patch size} 1\x16\x16 &{768}\x{16}\x{196}  \\ \hline
\multirow{2}{*}{masking} & tube mask & \multirow{2}{*}{  {768}\x{16}\x[{196}\x(1-{0.75})]} \\ 
{} & \emph{mask ratio} = 0.75 &{} \\ 
\shline
%\multirow{2}{*}{encoder} & \multirow{2}{*}{\(\left[\begin{array}{c}\text{MHA({768})}\\[-.1em] \text{MLP({3072})}\end{array}\right]\)$\times$12} \\ & \multirow{2}{*}{  {768}\x{8}\x[{196}\x(1-{0.75})]} \\
%{ }& { }& { } \\
\multirow{3}{*}{encoder} & \multirow{3}{*}{\(\left[\begin{array}{c}\text{MHA-T(768)} \\[-.1em] \text{MHA-S(768)} \\[-.1em] \text{MLP({3072})}\end{array}\right]\)$\times$12} & \multirow{3}{*}{  {768}\x{16}\x[{196}\x(1-{0.75})]} \\
%\shline
{ }& { }& { } \\
{ }& { }& { } \\ \hline

reshape & \emph{from} {{768}\x{16}\x({196}\x0.25)} \emph{to} {768}\x({16}\x{196}\x0.25) &  {768}\x({16}\x{196}\x0.25) \\ \hline

\multirow{2}{*}{regressor} & \multirow{2}{*}{\(\left[\begin{array}{c}\text{Cross-attn(768)} \\[-.1em] \text{MLP({3072})}\end{array}\right]\)$\times$4} & \multirow{2}{*}{  {768}\x{16}\x[{196}\x(0.75)]} \\
{ }& { }& { } \\ \hline

reshape & \emph{from} {768}\x({16}\x{196}\x0.25) \emph{to} {{768}\x{16}\x({196}\x0.25)} &  {{768}\x{16}\x({196}\x0.25)} \\ \hline

\multirow{3}{*}{appearance-decoder} & \multirow{3}{*}{\(\left[\begin{array}{c}\text{MHA-T(768)} \\[-.1em] \text{MHA-S(768)} \\[-.1em] \text{MLP({3072})}\end{array}\right]\)$\times$4} & \multirow{3}{*}{  {768}\x{16}\x[{196}\x(0.75)]} \\
%\shline
{ }& { }& { } \\
{ }& { }& { } \\ \hline

appearance-projector & Linear(16384) &   {16384}\x{16}\x[{196}\x(0.75)] \\ \hline
            
\multirow{3}{*}{motion-decoder} & \multirow{3}{*}{\(\left[\begin{array}{c}\text{MHA-T(768)} \\[-.1em] \text{MHA-S(768)} \\[-.1em] \text{MLP({3072})}\end{array}\right]\)$\times$2} & \multirow{3}{*}{  {768}\x{16}\x[{196}\x(0.75)]} \\
%\shline
{ }& { }& { } \\
{ }& { }& { } \\ \hline

motion-projector & Linear(768) &   {768}\x{16}\x[{196}\x(0.75)] \\ \hline

%\multirow{2}{*}{encoder} & \blockatt{768}{3072}{12} & \multirow{2}{*}{  {768}\x{8}\x[{196}\x(1-{0.75})]} \\

%& {} \\
%\shline
%{} \\
             
\end{tabular}
\vspace{.5em}
\caption{\textbf{Architectures details during the pre-training of our MAM$^2$.} ``MHA-T" and ``MHA-S" correspondingly denote the temporal and spatial multi-head self-attention. ``Cross-attn" denotes cross attention layer in a space-time agnostic manner. In particular, ``Cross-attn" takes the latent representation of visible patches with a shape of $[{768}\times({16}\times{196}\times0.25)]$ as the key and value representations and takes the mask queries witch a shape of $[{768}\times({16}\times{196}\times0.75)]$ as the query representations. We use masking ratio}
\label{tab:architecture} %\vspace{-2.5mm}
\end{table}
%##################################################################################################

\section{Tokenizer}
For the appearance prediction, we use the discrete variational autoencoder (dVAE) in VQGAN \citep{esser2021taming} to generate the visual tokens as the pre-training target. The dVAE acts as a visual tokenizer which transforms each video frame with the size of $224 \times 224$ into $14 \times 14$ visual tokens according to a pre-trained codebook where the vocabulary size of each visual token is $16384$. Note that we deprecate the decoder of the dVAE following \citep{bao2021beit, wang2022bevt}.

\section{Mask Sampling}
Fig.\ref{fig:mask} shows the \emph{cube} and \emph{tube} masking strategies. In particular, \emph{cube} mask is repeated from a 2-D block-wise mask on temporal dimension. And \emph{tube} mask is repeated from a 2-D random-wise mask on temporal dimension.

%\section{Pseudo-Code of VideoCAE}

%\section{Visualizations of attention maps for our VideoCAE pre-trained on Kinetics-400}

%You may include other additional sections here.

\end{document}